\documentclass[a4paper,conference]{IEEEtran}
\ifCLASSINFOpdf
\else
\fi
\usepackage{hyperref}
\usepackage{amsmath}
\usepackage{xcolor}
\usepackage{float}
\usepackage{graphicx}
\usepackage{multirow}

\begin{document}
%
\title{
A Few-shot Learning Approach for Historical {Ciphered} Manuscript Recognition
}

\author{\IEEEauthorblockN{Mohamed Ali Souibgui, Alicia Forn\'es}
\IEEEauthorblockA{Computer Vision Center\\
Computer Science Department\\
Universitat Aut\`{o}noma de Barcelona, Spain\\
Email: \{msouibgui,afornes\}@cvc.uab.es}
\and
\IEEEauthorblockN{Yousri Kessentini}
\IEEEauthorblockA{Digital Research Center of Sfax, 3021\\
MIRACL Laboratory, Sfax University\\ 
Sfax, Tunisia\\
Email: yousri.kessentini@crns.rnrt.tn}
\and
\IEEEauthorblockN{Crina Tudor}
\IEEEauthorblockA{Dept. of Linguistics and Philology\\
Uppsala University\\
Sweden\\
Email: crina.tudor@lingfil.uu.se}}


%


\maketitle

\begin{abstract}
Encoded (or ciphered) manuscripts are a special type of historical documents that contain encrypted text. The automatic recognition of this kind of documents is challenging because: 1) the cipher alphabet changes from one document to another, 2) there is a lack of annotated corpus for training and 3) touching symbols make the symbol segmentation difficult and complex. To overcome these difficulties, we propose a novel method for handwritten ciphers recognition based on few-shot object detection. Our method first detects all symbols of a given alphabet in a line image, and then a decoding step maps the symbol similarity  scores to the final sequence of transcribed symbols. By training on synthetic data, we show that the proposed architecture is able to recognize handwritten ciphers with unseen alphabets. In addition, if few labeled pages with the same alphabet are used for fine tuning, our method surpasses existing unsupervised and supervised HTR methods for ciphers recognition. 

\end{abstract}


%
\IEEEpeerreviewmaketitle

\section{Introduction}



\begin{figure}[h!]
    \centering
    \begin{tabular}{c}
      \includegraphics[width=0.95\columnwidth]{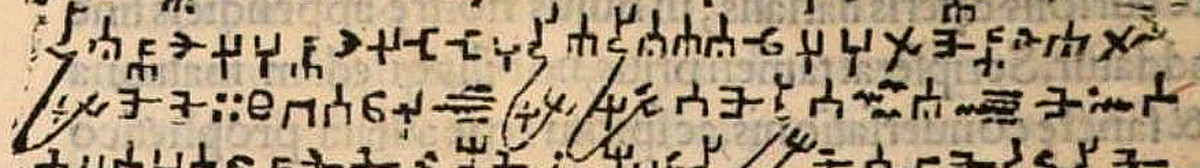}\\
      \includegraphics[width=0.95\columnwidth]{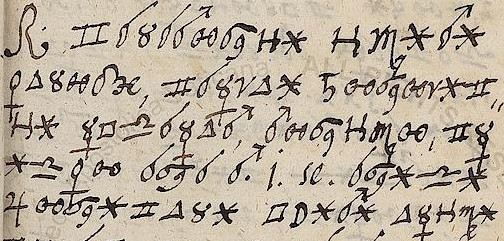}\\
      \includegraphics[width=0.95\columnwidth]{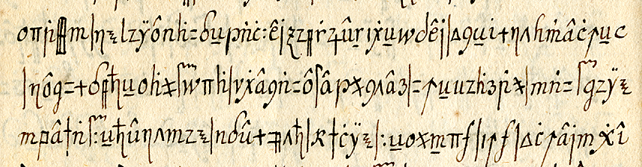}\\
    \end{tabular}
    \caption{Examples of handwritten ciphers dated from the 16th to the 18th century. Top: Devil cipher. Middle: Borg Cipher. Bottom: Copiale cipher.}
    \label{fig:ciphers}
\end{figure}

Historical documents residing in archives and libraries contain valuable information of our past societies. Despite the mass digitization campaigns for preserving cultural heritage, many historical documents remain unexploited unless they are properly transcribed and indexed.
One particularly interesting type of historical documents are ciphered manuscripts. Encoded (or ciphered) manuscripts are a specific type of historical documents existing in archives that contain secret messages or instructions. These documents used to be correspondence related to diplomatic, military, scientific or religious matters, among others. In order to hide their contents, the sender and receiver used to create their own secret method of writing, by transposing or substituting characters, special symbols, or by inventing a completely new alphabet of symbols. Some examples of historical ciphered manuscripts with invented alphabets are shown in Fig.\ref{fig:ciphers}.

Given the difficulties in the decryption of such manuscripts, some multi-disciplinar initiatives \cite{Megyesi-decrypt2020} have emerged to join the expertise in computer vision, computational linguistics, philology, criptoanalysis and history to make advances in historical cryptology. These joint efforts aim to ease the collection, transcription, decryption and contextualization of historical ciphered manuscripts in order to unlock their contents and make the secret information available for scholars in history, science, religion, etc.

The first step in the decryption process is the transcription, which consists in transforming the cipher document images into a machine encoded form (text). This task, called Handwritten Text Recognition (HTR), has been one of the most active fields in pattern recognition and document analysis \cite{Rejean2000}. With the recent advances in deep learning \cite{Gooffellow2016}, HTR systems can reach a very high performance \cite{Ciresan2011}, especially for modern documents with legible handwriting styles, known language, vocabulary and syntax. Contrary, the recognition of historical manuscripts is still challenging due to paper degradation, old vocabularies, uncommon handwriting styles, etc \cite{Stutzmann2018,Can2018,Rothacker2015}. To partially cope with these difficulties, context information is often incorporated through specific language models, dictionaries, etc. 

Obviously, the transcription of historical ciphered manuscripts is extremely difficult, due to three main reasons. First, and since the secret writing method aimed to encode the contents as much as possible, the underlying language was hidden, words were not separated by blank spaces, and no punctuation marks were used to separate sentences. Consequently, no language models nor dictionaries are available for helping the recognizer. But the greatest challenge appears when transcribing ciphers with unknown symbol alphabets. In such scenario, and since symbols frequently touch each other, even scholars doubt when identifying and segmenting each symbol in this new alphabet. This dilemma is explained by Sayre's paradox (characters cannot be properly recognized without being segmented and vice-versa).


In addition to the segmentation problem and the lack of context information, there is a third constraint to take into consideration: labeled data is barely available. Indeed, in many cases there is only one cipher document with the same symbols alphabet. This fact limits the use of deep learning architectures for transcribing ciphers.




Inspired by the recent advances in Few-shot learning \cite{snell2017,Sung_2018,Li_2019}, which has lately gained popularity in computer vision due to its ability to learn in limited data scenarios, in this work we explore the use of few-shot object detection for recognizing ciphers with unknown symbol alphabets. First, because very limited training examples are available, and second, because touching symbols of unknown alphabets makes unfeasible to use symbol segmentation methods. Moreover, a few-shot learning system does not need to be trained with the same testing classes. This enables the possibility of training with one symbol alphabet, and testing with another completely new and unseen alphabet. This is especially beneficial for recognizing ciphered documents of few pages containing unknown scripts without re-training the model. 

The main contribution of this work is the adaptation of few-shot object detection for transcription purposes (i.e. recognizing sequences of symbols). As far as we know, this is the first work based on few-shot learning for transcribing (ciphered) manuscripts. Our few-shot symbol detection model works at line level, because the segmentation into lines is rather easy and independent of the symbol alphabet. Hence, our system treats the alphabet of ciphers symbols as objects to be found in each handwritten line. Once all symbols are detected in the line image, we can use the spatial information of the bounding boxes to output the sequence of transcribed symbols. Here, we only need one or few (usually five as maximum) examples of each target symbol to be detected. The experimental results demonstrate the suitability of our approach. In addition, if we use very few real labeled data for fine tuning, our approach clearly outperforms existing supervised and unsupervised HTR approaches. 


The rest of this paper is as follows. We cover the related works to handwritten ciphers recognition in Section~\ref{related_works}. Then, we describe in detail the proposed approach in Section \ref{approach}. Afterwards, we present the experiments in Section \ref{experiments}. Finally, a brief conclusion is given in Section \ref{conclusion}.

\section{Related works}\label{related_works}

The recognition of handwritten ciphers can be seen as a particular case of Handwritten Text Recognition (HTR), where instead of a sequence of characters, the system must recognize a sequence of unknown symbols. Nowadays, most of the developed approaches for HTR focus on natural known scripts (Latin \cite{Kang2018}, Arabic \cite{Graves2009}, Chinese \cite{Zhang2017}, etc) and are based on deep learning architectures. Most models use Convolutional Neural Networks (CNN) and Recurrent Neural Networks (RNN), so they require a huge amount of annotated data and context information to learn in a supervised way the mapping function from the handwritten text image to the ground truth text class. As stated in the introduction, these models are inappropriate for cipher recognition for two main reasons. First, annotated ciphers are not available for training. Second, the alphabet of symbols usually changes from one ciphered document to another, which makes the building of a single HTR model even more complex. Although an attempt to transcribe handwritten ciphers using Multi-Dimensional Long Short-Term Memory Blocks Recurrent Neural Networks (MDLSTMs) \cite{Graves2009,Voigtlaender2016} was presented in \cite{Fornes2017}, this method needed labeled data from the same cipher alphabet to be trained. In any case, despite the efforts in annotating many cipher pages to train, the performance of this model decreased when applied to ciphers with invented alphabets \cite{baro2019}. 

Besides, some unsupervised methods were also proposed for ciphers recognition. In \cite{Yin2019}, the authors developed a method to segment the cipher documents into isolated symbols. Afterwards, a pretrained Siamese Neural Network (SNN) \cite{koch2015} was used to extract the features and cluster them with a Gaussian Mixture Model (GMM). The performance of this method was, however, moderate. The main problem was the difficulties in the symbol segmentation, since there was a big performance difference between the recognition results of the manually segmented symbols compared to the automatic segmented ones. Another unsupervised approach was proposed in \cite{baro2019}. After the symbol segmentation based on connected components, the K-means algorithm was used for clustering, followed by a label propagation to obtain the final transcription. The experiments showed that for an acceptable performance, a manual intervention was necessary during the selection and cleaning of clusters, mainly because of the difficulties in the segmentation of symbols. To summarize, these unsupervised methods could be useful to handle unknown alphabet symbol sets in ciphered manuscripts, but the segmentation of symbols is severely limiting their performance.

\section{Proposed approach} \label{approach}

In this section we describe how few-shot learning has been adapted to detect cipher symbols to transcribe encoded documents.

\begin{figure*}[h!]
    \centering
    \includegraphics[width=\linewidth]{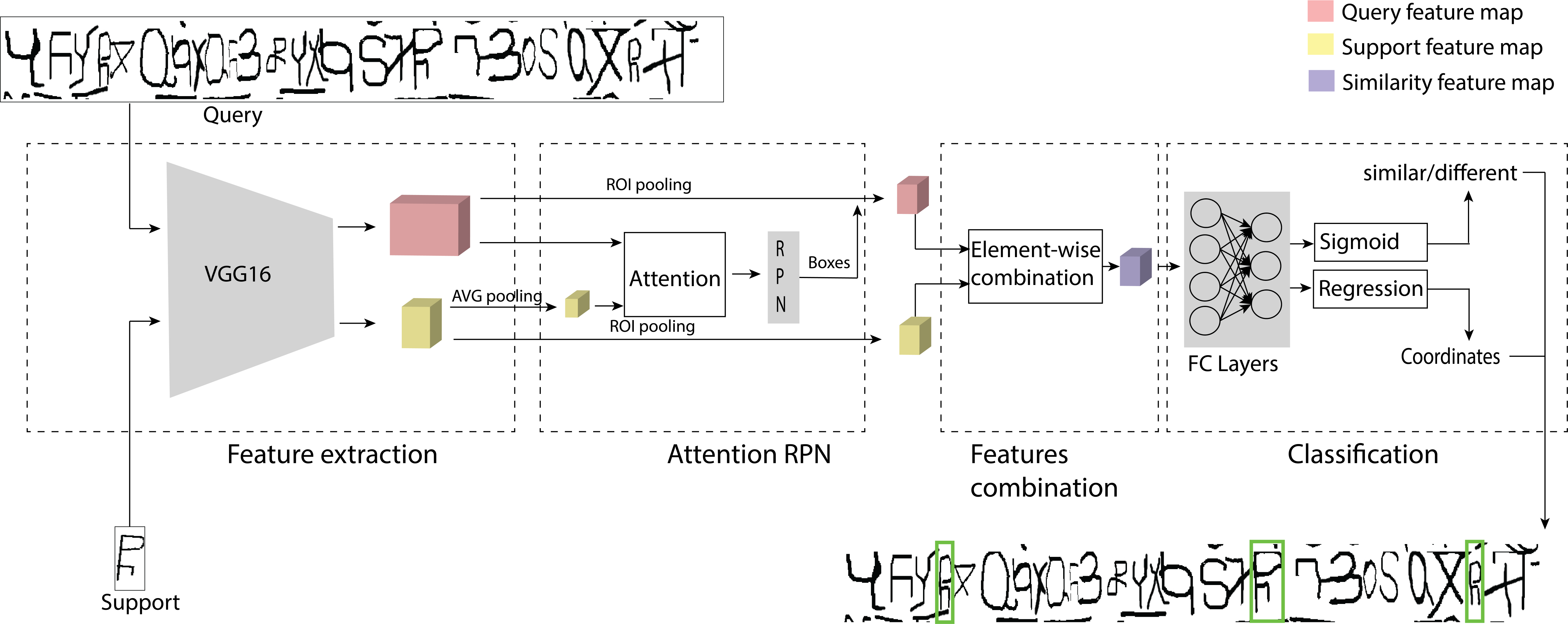}
    \caption{Our few-shot symbol detection architecture. The support and query image are the input of the VGG16 feature extractor. Two feature maps are obtained (with a depht of 512) and delivered to the attention Region Proposal Network (RPN). The RPN uses the two feature maps to compute a depth-wise cross correlation between them in order to obtain the attention feature map and generate  the region proposals. Next, Regions of Interest (ROI) pooling is applied on the support and the proposed regions features   to obtain similar shapes. Afterwards, they are combined and passed to the classifier to predict the bounding box coordinates together with the assigned class (similar or different from the support image). In the output, we show in green color the bounding boxes of those retrieved areas that contain a symbol similar to the support image.}
    \label{fig:Architecture}
\end{figure*}

\subsection{Background: Few-shot Learning}

Few Shot Learning is a subarea of machine learning that consists in feeding a learning model with very few labeled data (e.g. a classifier is only given few examples of each category to make a decision). Formally, in the few shot setting scenario, our dataset is divided in a training set and a test set with disjoint label sets (i.e. different classes). Then, the test set is further divided into pairs of a query, whose label is unknown, and a support set with known labels. Thus, a few-shot model must adapt to the few labels provided by the support set in order to classify the given query. Note that these labels (classes) have never been seen during training. Therefore, the classifier is learning a matching (or similarity) function between the query and the support images so that it can be adapted to the new classification problem during test time. If the support set contains $K$ labeled examples for each of the $C$ unique classes, the few-shot problem is called $C$-way $K$-shot. 

\subsection{Problem Definition}

Few Shot Learning has been typically applied to classification tasks. Lately, the application of few-shot learning to other topics, such as object detection is recently getting the attention of the computer vision community  \cite{Fan2019,Chen2018,B-kang2019,Karlinsky2019}.
Inspired by these few-shot object detection approaches, we aim to adapt those techniques to recognize the encoded manuscripts by detecting the ciphered symbols at line level. So, given an input handwritten line and one or few cropped examples of a desired symbol class, the model should be able to detect all symbols belonging to this class within the line. Since the model has not been trained on the same real ciphered dataset, our problem can be defined as few-shot symbol detection, following the trend of the so-called few-shot object detection. In this case, the handwritten lines are the query images and the ciphered symbols are the support images. For each cipher, if the symbol set (or the support images) contains $N$ classes and we are providing $K$ examples from of each symbol class, the problem is considered $N$-way $K$-shot detection. 

\subsection{Few-shot Symbol Detection Architecture}

Our proposed segmentation-free few-shot symbol detection architecture is inspired from \cite{Fan2019}. Note that symbol detection might be a more complex process compared to the object detection in scene images, because of the high similarity between the handwritten symbols. Moreover, since our goal is handwritten ciphers recognition instead of simply object detection, our approach extends the detection method to transform the retrieved objects as a sequence of symbols (the transcribed text line). Note also that we used a different backbone and similarity detection function compared to the original method, which was more suitable to our problem. We designed the model to be as a Siamese version of the Faster R-CNN detector \cite{Ren2015}. It  learns a similarity function between the support images and the found bounding boxes on the query image. After the detection, a decoding algorithm is used to read the text by combining the spatial information of the detected symbols. {So, to use the proposed symbol detector in ciphered manuscript recognition, the user should provide the number of classes to transcribe and one or few examples of each class.}

The overall architecture of our proposed system is, hence,  presented in Fig.~\ref{fig:Architecture}. It contains four main steps: feature extraction, regions proposing, features combination and classification. The different stages are described next:

\subsubsection{Feature Extraction}
The first step consists in extracting the features representations for the query line and the support images (corresponding to examples of the symbol alphabet). Thus, when the model receives the query and the support as input, it propagates them through a VGG16 \cite{Simonyan2015} backbone to obtain the query features and the support features, respectively. The backbone weights were not pretrained, and they are shared because we are using a Siamese architecture.

\subsubsection{Regions proposal} 
The query and the support features maps are passed to the Region Proposal Network (RPN), which includes an attention mechanism. First, the attention RPN applies a repetitive average pooling on the support features map until  obtaining a $1 \times 1 \times C$ shape, where $C$ is the number of channels of the features. This pooled result is used to compute the following element-wise multiplication to get the attention features map as a result. Denoting by $S \in t^{1,1,C}$ the support feature map after the pooling and by 
$Q \in t^{W,H,C}$ the query features. The attention feature map $A \in t^{W,H,C}$ is defined by:
\begin{equation}
   A_{w,h,c} =  S_{1,1,c} \cdot Q_{w,h,c}
\end{equation}
Where,  $w \in \{1,...,W\}$, $h \in \{1,...,H\}$ and $c \in \{1,...,C\}$.
$A$ is used to propose the regions.  The attention RPN was empirically showed in  \cite{Fan2019} to be more beneficial than the standard one, since it helps to provide more relevant regions boxes for the next steps by taking the support features into account.

\subsubsection{Feature combination}

At this step, the Region of Interest (ROI) pooling is applied on both the RPN proposals and the supports feature maps to obtain two tensors with the same size ($7 \times 7 \times C$). In our case, we are using a well cropped images in the support set. Thus, all the support image is considered as a GT bounding box before ROI pooling its feature map. Then, we use a simple subtraction to combine the two ROI pooled feature maps. We are aware that, at this step, many other relation detection functions could be used for the same purpose \cite{Fan2019} (e.g. addition, concatenation, multiplication, etc). The combined feature map is then passed to a predictor to get the bounding boxes.

\subsubsection{Classification}

The predictor consists in fully connected layers with two final outputs. First, a classifier with a sigmoid activation function is used to decide whether the class of each proposed region is similar (1) or different (0) from the support symbol. The second output consists in the spatial information of the final bounding boxes (i.e. their coordinates). For this purpose, a regression is performed to fit the correct bounding box for each detected symbol. 

\subsection{Transcription}

\begin{figure}[h!]
    \centering
    \includegraphics[width=\linewidth]{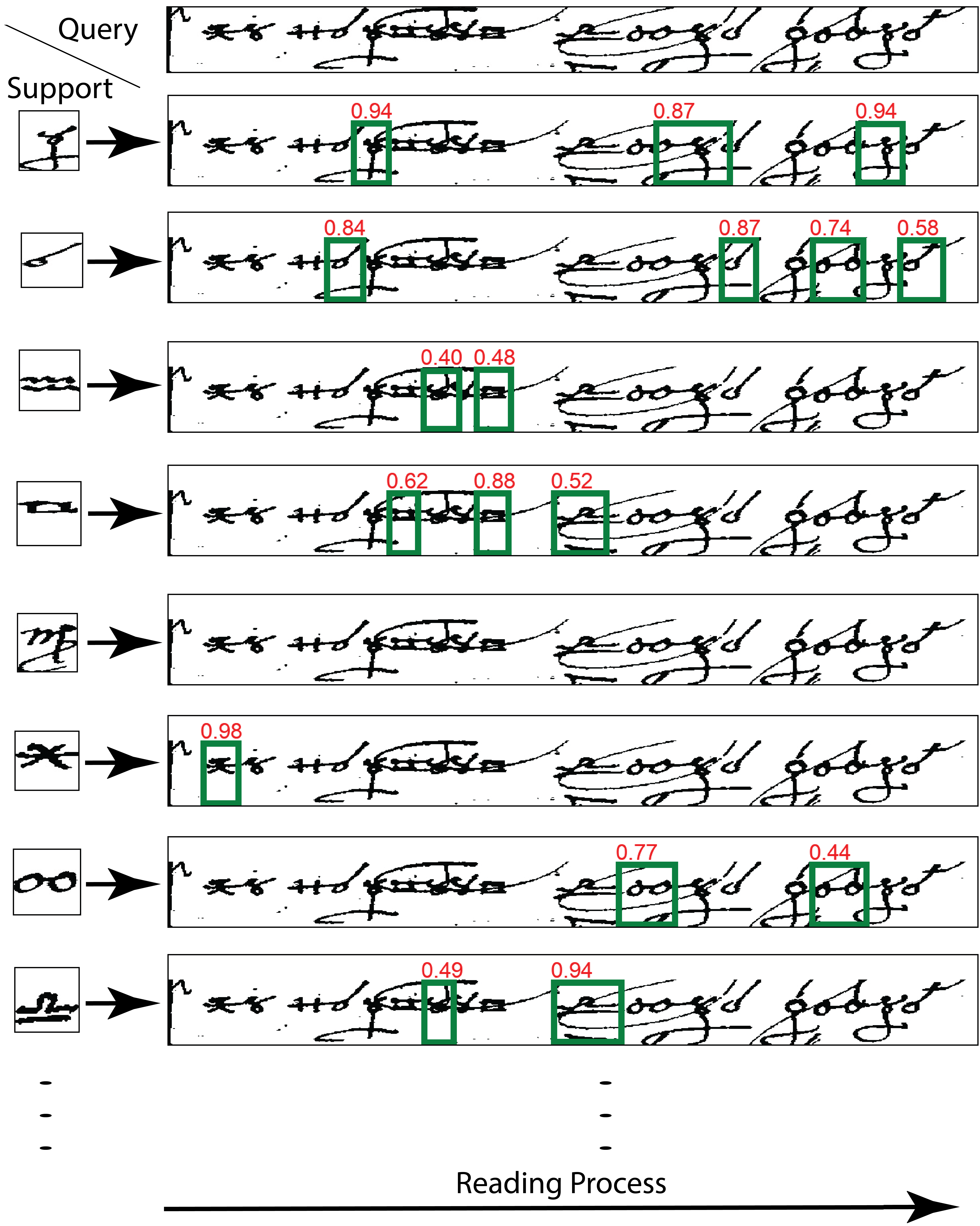}
    \caption{Detected Borg symbols in a handwritten cipher line. We only show in green color the retrieved boxes with a similarity score of minimum 0.4.}
    \label{fig:borg_boxes}
\end{figure}

Once the few-shot symbol detector has been applied, a decoding algorithm is needed to obtain the final transcribed ciphered text. Indeed, by providing as input one image line as a query and the cipher alphabet as support images (one at each time) to the few-shot symbol detector, we obtain the potential bounding boxes coordinates for all the support symbols. Figure~\ref{fig:borg_boxes} illustrates this as a two-dimensional table, since the detection at line level only requires the $y$-axis to represent the detected boxes (so, it starts at column $y_1$ and ends at column $y_2$). The symbols alphabet to be searched are shown in the column "Support".  For each symbol, the retrieved bounding boxes in the query line image are shown in green, and the assigned score (in red color, with values between 0 and 1) indicates the similarity degree. 

Until now, we have kept all the retrieved candidate bounding boxes, which means that, for each column in the query line, more than one symbol alphabet can be retrieved. This is  a frequent case, because some symbol classes in an alphabet can have a similar appearance. So, in order to obtain the final transcription, the decoding algorithm traverses the cipher line from left to right, deciding, at each column, which will be the final transcribed symbol class among the candidate symbols. For this purpose, for each time step, the symbol with the maximum similarity score will be transcribed, discarding the other candidate symbols. While traversing the line (from left to right), if another bounding box with a higher score is found, then we assume that we are now transcribing the next symbol within the line. This process is repeated until the end of the cipher line is reached. Obviously, a symbol is only transcribed if its bounding box is not interrupted (overlapped) by another symbol with a maximum value nearby (in our case we used 15 pixels as a threshold).

Despite its simplicity, this decoding algorithm turns out to be effective for transcribing the sequence of cipher symbols, especially if we take into account that language models (e.g. n-grams) nor dictionaries are available. Of course, if labelled data or context information would be available, the decoding could be highly enhanced by using, for example, beam search or sequence-to-sequence deep learning models. We finally note that we are learning the full bounding boxes rather than the $y$-axis coordinates (which could be simpler) so that we can easily scale our approach to detect symbols at paragraph or page level.

\section{Experiments and results} \label{experiments}

In this section we describe how the training settings and  data, the experimental evaluation using real cipher manuscripts, and the comparisons with existing methods in the literature.

\subsection{Training data}

Since our approach is based on few shot learning, it can be trained with different classes than those of the testing set. In our case, we use the Omniglot dataset \cite{Lake2015}, a widely used dataset designed for developing more "human-like" learning algorithms. This dataset, which has been lately used to test few(one)-shot approaches, contains 1623 different handwritten characters categories from 50 different alphabets, with only 20 examples per category, each one drawn by a different writer.

\begin{figure}[h!]
\centering
\begin{tabular}{c}
  \includegraphics[width=0.95\columnwidth]{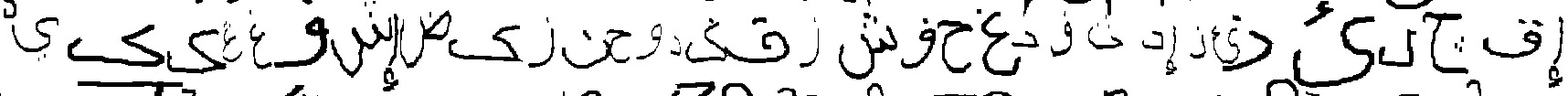}\\
  \includegraphics[width=0.97\columnwidth]{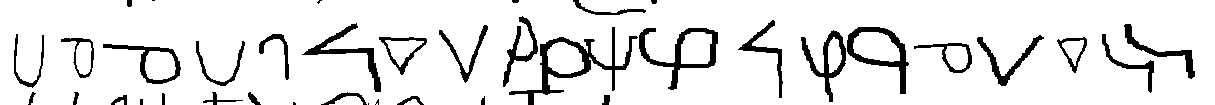}\\
  \includegraphics[width=0.97\columnwidth]{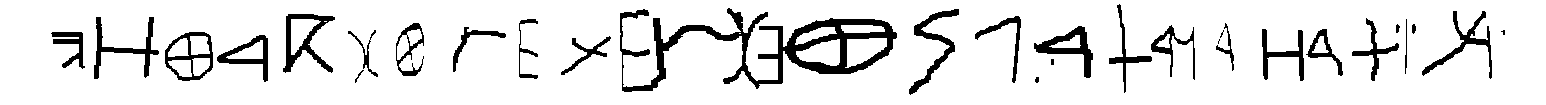}\\
  \includegraphics[width=0.97\columnwidth]{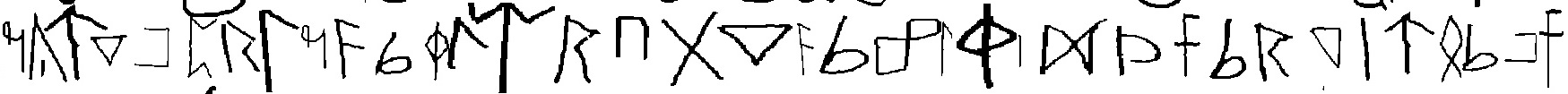}\\
  \includegraphics[width=0.97\columnwidth]{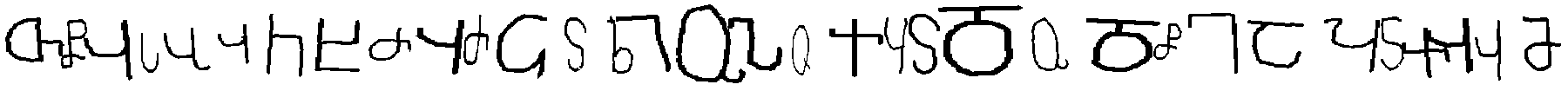}\\
  \includegraphics[width=0.97\columnwidth]{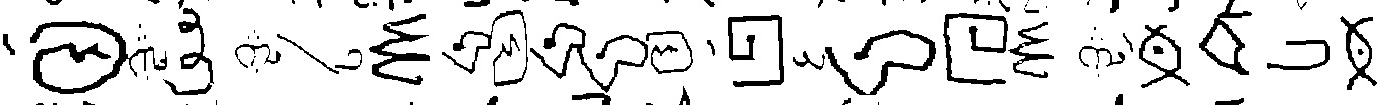}\\
\end{tabular}
\caption{Examples of generated synthetic lines using Omniglot symbols.}
\label{fig:lines}
\end{figure}

So, for training our model, we create synthetic query lines that simulate handwritten ciphered lines. For this, we take the "images\_background" Omniglot subset (from 30 different alphabets), which correspond to 964 different symbol classes. It must be noted that a high amount of classes usually benefit few-shot training, since the generalization ability to unseen classes increases if the model can see many different categories. Given that the number of examples for each class should be low, we use the first 7 samples of each class to create the synthetic query lines, meanwhile the last 10 samples are used to randomly choose the desired support images for each class. They are first passed through a pre-processing step, where some random transformations are applied (e.g. resizing, rotation, dilation, etc). Thus, we have created 2000 lines by placing these symbols as a sequence, with a random distance between them and a quite high probability of overlapping symbols, either horizontally or vertically (emulating touching cipher lines). We keep the randomly chosen spatial information as the ground truth bounding boxes of the resulting images. Note that each line can contain from 5 to 50 symbols. Some examples are presented in Fig.~\ref{fig:lines} for further illustration. During training, the bounding boxes that contain similar symbols to the support image are labeled 1, and all the different symbols are labeled 0 (background).

\subsection{Testing datasets}

\begin{figure}[h!]
    \centering
    \begin{tabular}{c}
      \includegraphics[width=0.95\columnwidth]{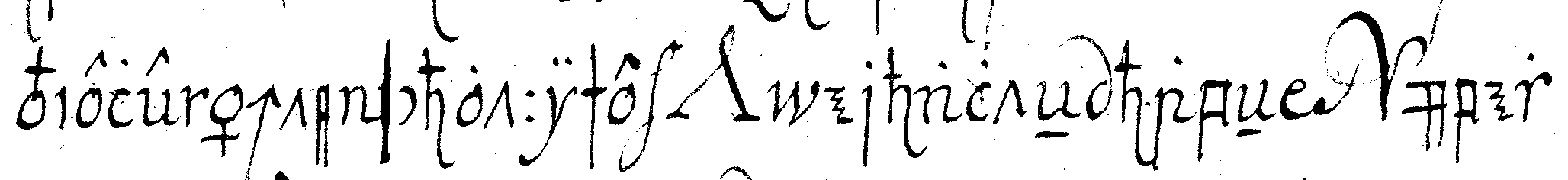}\\
      \includegraphics[width=0.95\columnwidth]{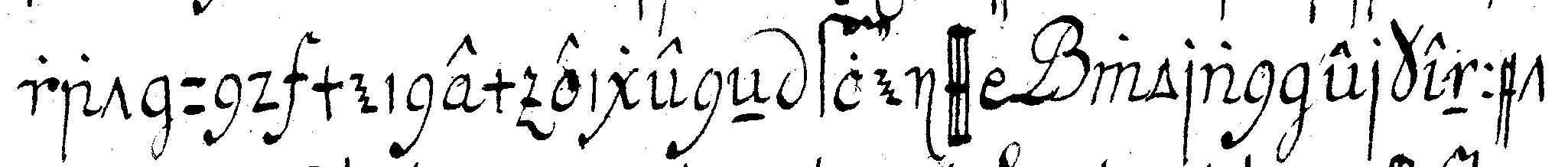}\\
      \includegraphics[width=0.95\columnwidth]{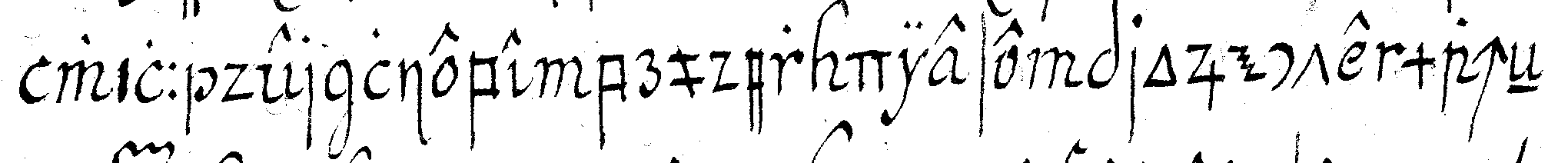}\\
      \includegraphics[width=0.95\columnwidth]{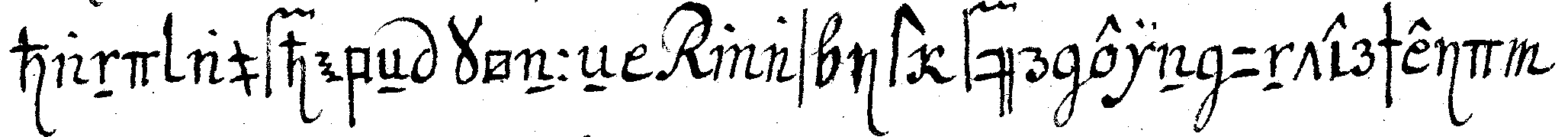}\\
    \end{tabular}
    \caption{Four query line images from the Copiale cipher.}
    \label{fig:copiale}
\end{figure}

Our model, once trained with synthetic data, will be evaluated on two real historical encoded manuscripts taken from the DECODE \cite{Megyesi2019} database: Copiale\footnote{\url{https://cl.lingfil.uu.se/~bea/copiale/}} and Borg\footnote{\url{https://cl.lingfil.uu.se/~bea/borg/}} ciphers.   
The Copiale cipher is a 105 pages manuscript dated back to 1760-1780, containing around 75,000 symbols. The cipher alphabet has 90 different symbol classes, which is composed of Roman and Greek letters, diacritics and abstract symbols. For a fair comparison, we have used the same 52 pages for testing that have been used in \cite{baro2019}. The chosen pages have been binarized \cite{Sauvola2000} and segmented into query lines using projections. An example of the obtained lines is showed in Fig~\ref{fig:copiale}. We have cropped the support images from different pages. The choice of the support images was random during testing.

\begin{figure}[h!]
    \centering
    \begin{tabular}{c}
      \includegraphics[width=0.9\columnwidth]{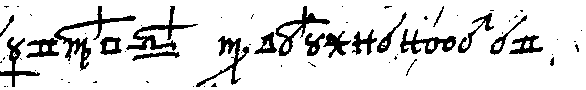}\\
      \includegraphics[width=0.9\columnwidth]{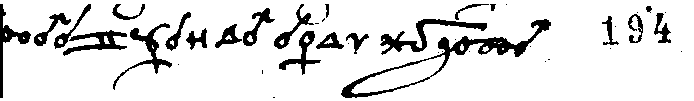}\\
      \includegraphics[width=0.9\columnwidth]{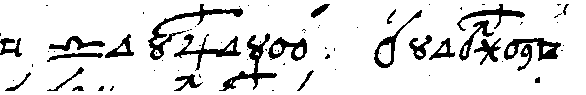}\\
      \includegraphics[width=0.9\columnwidth]{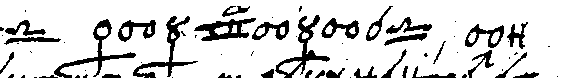}\\
    \end{tabular}
    \caption{Four query lines images from the Borg cipher. Note that in this manuscript, there are frequent touching symbols, even from different lines.}
    \label{fig:borg}
\end{figure}

The Borg cipher is a 408 pages manuscript from the 17th century. Its alphabet has 34 different symbols composed of abstract, esoteric symbols, Roman numbers, and some diacritics. Some examples of the Borg lines are shown in Fig.~\ref{fig:borg}. Note that this cipher is much harder to be processed compared to Copiale, mainly because of the frequent symbol overlapping (between consecutive symbols and also between symbols from to different lines). Following \cite{baro2019}, we have used the same 16 pages from this manuscript for testing. Same as Copiale, a pre-processing step (binarization plus projections) has been applied for segmenting the pages into lines. Information about the specific pages used in this experiments are available\footnote{\url{https://cl.lingfil.uu.se/decode/publ.html}}.

\subsection{Training settings and metrics}


We compare our approach with other existing works for ciphers recognition, categorized in two main classes: the unsupervised methods \cite{baro2019} and \cite{Yin2019}, which were designed, same as ours, for handwritten ciphers with invented alphabets, and the supervised HTR method based on MDLSTM applied to numerical ciphers \cite{Fornes2017} that was also applied to Borg and Copiale ciphers in \cite{baro2019}. During the experiments, we tried different settings to analyze the performance of our model in depth. In our method, we vary the number of shots (i.e. the number of examples per class in the support images) and the confidence threshold. The confidence threshold means that we only transcribe the predicted symbol if its similarity score is higher than a given threshold (we used 0.4, 0.6 or 0.8 in our experiments). As explained in \cite{baro2019}, the use of a confidence threshold is necessary for ciphered handwritten recognition because whenever the similarity score is low, it is better to ask for human intervention rather than making a wrong prediction, which would lead to error propagation during the deciphering process. Indeed, few errors in the transcription can highly affect the cryptanalysis (i.e. finding the deciphering key). 

As evaluation metric, we follow  \cite{baro2019} in using the Symbol Error Rate (SER)  which is based on the Character Error Rate (CER) for text recognition. Formally, $SER = \frac{S+D+I}{N}$, where S is the number of substitutions, D of deletions, I of insertions and N the ground-truth's length. So, the lower value, the better.

We also compute the percentage of missing symbols, which indicates the amount of symbols that remain untranscribed, so the user must manually transcribe them. The percentage of missing symbols varies depending on the confidence threshold, so a high confidence threshold means that only those symbols with a high similarity score will be automatically transcribed. Obviously, if the confidence threshold increases, the SER decreases, but the percentage of missing symbols increases, which implies a higher user intervention for getting the final transcription. Thus, the ideal method should find a trade-off between the SER and the percentage of missing symbols.

\subsection{Results}


\begin{table*}[h!]
\caption{Comparison with other existing works for ciphers recognition. The Confidence, Symbol Error Rate (SER) and Missing symbols range from [0-1]. The lower SER and Missing symbols values, the better performance.}
\label{tab:results}
\begin{center}
\begin{tabular}{|c|c|c|c|c|c|c|c|c|}
\hline
 \multirow{2}{4em}{\textbf{Method}} & \multirow{2}{2em}{\textbf{Shots}} & \multirow{2}{*}{\textbf{Confidence}} & \multicolumn{3}{|c|}{\textbf{Copiale cipher}}& \multicolumn{3}{|c|}{\textbf{Borg cipher}} \\
 \cline{4-9}
 {}&{}&{}&\textbf{Labeled pages}&\textbf{SER}&\textbf{Missing symbols}&\textbf{Labeled pages}&\textbf{SER}&\textbf{Missing symbols}\\
 \hline
 SNN+GMM \cite{Yin2019}&-&-&0&0.440 &-&0&0.570&-\\
 \cite{Yin2019}+Manual segmentation &-&-&0&0.370&-&0&0.220&-\\
 \hline
 \hline
 Clusters+LabelProp \cite{baro2019} &-&0.4&0&0.444&0.031&0&0.542&0.377\\
 \cite{baro2019}+Manual cluster select. &-&0.4&0&0.201&0.010&0&0.522&0.173\\
 \hline
 \multirow{4}{*}{MDLSTM \cite{Fornes2017}} &-&\multirow{4}{*}{0.4}&25& 0.131&0.008&7&0.715&0.154\\
 & -&{}&34 & 0.120&0.009&9&0.662&0.069\\
 &-&{}& 42& 0.084&0.006&11&0.693&0.061\\
 &-&{}& 51& \textbf{0.075}&0.003&14&0.556&0.035\\
\hline
\multirow{6}{*}{Ours}&1&\multirow{6}{*}{0.4}&0&0.462&0.041&0&0.720&0.013\\
&5&{}&0&0.392&0.029&0&0.534&0.096\\
&1&{}&2&0.112&0.010&2&0.236&0.010\\
&5&{}&2&0.114&0.005&2&0.236&0.008\\
&1&{}&5&0.111&0.009&5&0.220&0.020\\
&5&{}&5&0.115&0.002&5&\textbf{0.210}&0.012\\
\hline
\hline

 Clusters+LabelProp \cite{baro2019} &-&0.6&0&0.365&0.073&0&0.445&0.547 \\
 \cite{baro2019}+Manual cluster select. &-&0.6&0&0.173&0.024&0&0.464&0.282\\
 \hline
 \multirow{4}{*}{MDLSTM\cite{Fornes2017}} &-&\multirow{4}{*}{0.6}&25& 0.113&0.068&7&0.554&0.399\\
 & -&{}&34 &0.109&0.076&9&0.523&0.280\\
 &-&{}& 42&0.078&0.048&11&0.551&0.269\\
 &-&{}& 51&\textbf{0.074}&0.038&14&0.450&0.212\\
\hline
\multirow{6}{*}{Ours}&1&\multirow{6}{*}{0.6}&0&0.388&0.085&0&0.661&0.161\\
&5&{}&0&0.336&0.053&0&0.513&0.115\\
&1&{}&2&0.100&0.023&2&0.206&0.029\\
&5&{}&2&0.103&0.009&2&0.211&0.016\\
&1&{}&5&0.094&0.020&5&0.197&0.030\\
&5&{}&5&0.093&0.009&5&\textbf{0.193}&0.017\\
\hline
\hline

 Clusters+LabelProp \cite{baro2019} &-&0.8&0&0.264&0.138&0&0.377&0.713 \\
 \cite{baro2019}+Manual cluster select. &-&0.8&0&0.144&0.045&0&0.418&0.385 \\
 \hline
 \multirow{4}{*}{MDLSTM\cite{Fornes2017}} &-&\multirow{4}{*}{0.8}&25&0.110&0.154&7&0.546&0.513\\
 & -&{}&34 &0.111&0.170&9&0.437&0.435\\
 &-&{}& 42&0.087&0.116&11&0.465&0.421 \\
 &-&{}& 51&0.081&0.098&14&0.365&0.365 \\
\hline
\multirow{6}{*}{Ours}&1&\multirow{6}{*}{0.8}&0&0.294&0.241&0&0.396&0.368\\
&5&{}&0&0.269&0.142&0&0.361&0.253\\
&1&{}&2&0.072&0.081&2&0.175&0.082\\
&5&{}&2&0.075&0.042&2&0.188&0.049\\
&1&{}&5&\textbf{0.067}&0.082&5&\textbf{0.168}&0.084\\
&5&{}&5&0.069&0.034&5&0.173&0.043\\
\hline

\end{tabular}
\end{center}
\end{table*}

The different existing methods have used several testing settings. In the unsupervised methods, there is a first scenario with no manual intervention (so, the method is fully automatic), and a second scenario with some user intervention during testing (basically, a manual segmentation \cite{Yin2019} and cleaning of clusters \cite{baro2019}). In the supervised MDLSTM \cite{Fornes2017}, the number of labeled pages to train the system varies, which implicitly measures the human effort (i.e. user intervention) to provide labelled data to train the HTR system. In the comparatives, the results of the different settings are reported. 

Similarly, at test time, our proposed architecture has been evaluated in two different scenarios:

\begin{enumerate}
    \item Without retraining: In this scenario, we directly use the trained model on the synthetic Omniglot data to transcribe the real ciphers following the few-shot setting. This means that our  approach must adapt and generalize to other unseen cipher alphabets classes.
    
    \item With retraining (fine tuning): Despite our attempts of creating synthetic data that emulates real ciphers, there are many visual differences between the datasets, especially between the synthetic data and the Borg cipher. Consequently, the model could barely adapt and generalize to the new classes and domains at the same time. Therefore, in addition to the pure few-shot scenario described above (no retraining), we also explore a second scenario, where we relax this constraint and allow some retraining for fine-tuning, as done in \cite{Chen2018} for few-shot object detection. Thus, after training with Omniglot data, our model is retrained using a very small amount of real data (examples of the real cipher symbols) for fine tuning. We used for retraining 2 or 5 labeled pages. Each page contains approximately 17 lines. 
\end{enumerate}

Obviously, from the point of view of the user, the first scenario is better because the model handles unseen alphabets without needing real label data to retrain. However, from the point of view of the automatic recognition, the second scenario is better because if the data distributions between the synthetic lines and the real ciphered manuscripts are very different, the result is a poor system performance (i.e. a high SER).

All the experimental results are presented in Table~\ref{tab:results}. As expected, a high confidence threshold implies a higher percentage of missing symbols. We also note that more shots (5-shots instead of 1-shot) usually improve the SER (especially when there are different handwriting styles). Otherwise, it improves the missing symbols rate, which is also beneficial. {However, providing more shots implies more user effort (it may also require checking several pages to label the symbols), so users might prefer the 1 shot option.}  

Given the high amount of touching symbols in the Borg cipher, the transcription performance of the Copiale cipher is significantly better. Obviously, the use of 2 or 5 labelled pages for fine tuning (our second scenario) boosts the performance, obtaining the best transcription accuracies (the SER is 0.067 in Copiale, and 0.168 in Borg).
This performance implies a rather low user intervention. Indeed, the user effort in labelling 2-5 pages is significantly lower than labelling 14-51 pages for training the MDLSTM model.

In the first scenario (without retraining), our few-shot model outperforms all the unsupervised (clustering) methods, especially when there is no user intervention. Our approach is rather competitive compared to the supervised MDLSTM one, especially when it is trained on a small amount of data (in Borg for example, our model with 5 shots is close to MDLSTM in many cases). Obviously, the incorporation of many labelled pages to train the MDLSTM is highly beneficial, but it implies a high human effort. 

In the second scenario (fine tuning), our model considerably improves. As we said before, it outperforms all the existing approaches, even when using only 2 labeled pages for retraining. Our method also obtains the best results in the Copiale cipher, giving the best result with the confidence threshold of 0.8. Of course, if we retrain with 5 pages, the model still improves. However, given the slight performance improvement, and considering the human effort (labelling 5 pages instead of 2), we can conclude that fine tuning with 2 labelled pages is the most suitable scenario.

\section{Conclusion} \label{conclusion}

In this paper, we have presented a few-shot symbol detection model for the particular case of historical encoded manuscripts recognition. We showed throught the experimental results the suitability of our approach, outperforming the unsupervised methods without the need of labeled data. In addition, if we fine-tune with few real data, our model achieves the state of the art for Borg and Copiale recognition, outperforming a supervised approach based on MDLSTM. Since there is very few work in the literature that addresses the recognition of historical encoded manuscripts with unknown alphabets, we hope that our novel few-shot learning based recognition method can serve as a baseline and foster the research in this topic.

As future work, we will investigate the recognition of ciphers at page level, avoiding the segmentation into lines. We will also test different settings in the model components, for instance: Finding the suitable feature combination method, and  extraction as well with the possibility of pretraining the backbone on the right data. Furthermore, we plan to explore unsupervised methods to automatically suggest the alphabet of symbols, so that one can easily obtain the support images for the few-shot detection model. Finally, we aim to improve the generation of synthetic lines to better resemble the real ciphers. In this way, it will not be necessary to retrain the system with real data for obtaining a good recognition performance.

\section*{Acknowledgment}

The authors would like to thank all the volunteers that have been labeling the Borg and Copiale ciphers. This work has been partially supported by the Swedish Research Council (grant 2018-06074, DECRYPT), the Spanish project RTI2018-095645-B-C21, the Ramon y Cajal Fellowship RYC-2014-16831 and the CERCA Program / Generalitat de Catalunya.



%

\bibliographystyle{ieeetr}
\bibliography{main}




\end{document}